# Approximate Learning in Complex Dynamic Bayesian Networks

**R. Settimi**
Department of Statistics,
University of Warwick,
Coventry, CV4 7AL, U.K.
e-mail: R.Settimi@warwick.ac.uk

**J. Q. Smith**
Department of Statistics,
University of Warwick,
Coventry, CV4 7AL, U.K.
e-mail: J.Q.Smith@warwick.ac.uk

**A. S. Gargoum**
Department of Statistics,
College of Business and Economics
U. A. E. University
Al-Ain - United Arab Emirates

## Abstract

In this paper we extend the work of Smith and Papamichail (1999) and present fast approximate Bayesian algorithms for learning in complex scenarios where at any time frame, the relationships between explanatory state space variables can be described by a Bayesian network that evolve dynamically over time and the observations taken are not necessarily Gaussian. It uses recent developments in approximate Bayesian forecasting methods in combination with more familiar Gaussian propagation algorithms on junction trees. The procedure for learning state parameters from data is given explicitly for common sampling distributions and the methodology is illustrated through a real application. The efficiency of the dynamic approximation is explored by using the Hellinger divergence measure and theoretical bounds for the efficacy of such a procedure are discussed.

**Keywords**: Dynamic Generalised Linear Models, Dynamic Bayesian networks, Hellinger distance, junction trees.

## 1 INTRODUCTION

In the last ten years, Bayesian probabilistic networks have been largely applied to complicated high-dimensional problems, where the relationships among several variables are uncertain. When the dependence structure in the problem domain remains fixed, Bayesian inference and efficient algorithms for structural learning and data propagation are well-established and widely used in the Statistical and Artificial Intelligence community.

However in dynamic situations, where typically the domain together with the interdependencies among the variables evolve with time, Bayesian inference and updating algorithms can become progressively hard (see Boyen and Koller (1998) for an interesting discussion of these issues). To model such scenarios Bayesian probabilistic networks need to be defined over state spaces, i.e. the sets of uncertain quantities or parameters defining the stochastic process, which vary over time.

The first problem faced in dynamic contexts is that the state space will tend to become rather complicated and may grow enormously with the passage of time. New variables may be added to the problem and new conditional independencies are created or old ones disappear. Moreover, in the applications we have in mind, where the production of forecast distributions at each time step is critically important, observations will arrive sequentially and the system will need to be updated sequentially after the arrival of each new data and be ready to receive new information that may arise.

This scenario occurs in many fields. In clinical investigations data associated with symptoms are monitored over time. The evidence accumulates as the disease develops, so different symptoms arise and new data need to be observed for each patient. In marketing research for instance we need to predict the sales of new products in competing markets where new competitors are constantly being introduced (see Smith and Papamichail, 1999). In the analysis of the fluctuations and behaviour of financial stocks quick forecasts and predictions are usually demanded in an environment where causal variables are constantly changing. In section 5 of this paper we shall illustrate our data propagation technique through an environmental application which involves a complex dynamic process associated to the radioactive gaseous mass release in the event of a nuclear accident (Smith et al. 1995).

Because of the difficulties mentioned above, standard algebraic updating procedures for Bayesian networks that are extremely efficient in the static case may become unfeasible when applied to the dynamic context. On the other hand, numerical Monte Carlo methods tend to be slow and are fraught with complexity difficulties compared to their static counterparts even in much simpler domains (see Shepard and Pitt, 1999).

The dependence structure in the dynamic problems can be represented by using graphical models whose structure changes dynamically, accordingly to the evolution of the state space. Following Smith and Papamichail (1999), we can exploit such a representation to develop approximate algorithms of propagation and probability updating for non Gaussian dynamic systems that use well-established results for non-dynamic situations. These algorithms are based on a generalisation of a time series method first introduced by West and Harrison (1997) and adopted in Smith (1992). Such methods are widely used in the simpler time series setting, for a recent review see Durbin and Koopman (1999). Although these procedures are approximate they are algebraic and so extremely efficient. In section 4 and 5 we show how the exactness of the approximation can be measured formally. In the application we have investigated, the approximations are extremely good. Finally note that even in applications where there are no time restrains and accurate numerical routines can be run for long periods of time, this approximate process is useful both to set the initial distributional inputs and to monitor the performance of the numerical algorithms which are often unstable.

We conclude this introduction by reviewing a little background material from Bayesian graphical modelling in the time dependent perspective. A typical assumption in dynamic modelling is that domains are assumed to be Markov, that is predictions about future states depend only on the current states and therefore to retrieve information at a certain point of time $t$ we just need to know the state of the system at time $t-1$. We assume also that the probability distribution over the states $\Theta_T = \{\theta_0, ..., \theta_T\}$ specifying the stochastic system till time $T$, is represented as a directed acyclic graph (DAG) $\mathcal{G}_T$, whose nodes are the parameters in $\Theta_T$ and that represent the conditional independence structure implicit in the problem at each time $T$. The graphical model $\mathcal{G}_T$ not only provides a compact representation of the joint probability distribution of the process but also allows us to utilise results and efficient algorithms that are developed for static Bayesian networks.

In particular it is possible to employ the popular propagation algorithms defined over junction trees (see e.g. Spiegelhalter et al. 1993, Jensen et al., 1994). A junction tree is defined as an undirected decomposable graph that can be constructed from a DAG $\mathcal{G}$ as follows:

1. form the decomposable graph $\bar{\mathcal{G}}$ by linking all the nodes in $\mathcal{G}$ that have the same child node, i.e. have edges directed to the same node;

2. identify a sequence of cliques in $\bar{\mathcal{G}}$, where a clique is a maximally connected subset of nodes, that satisfies the running intersection property, i.e. given the sequence $C[1], ...C[m]$, for any $i, 2 \leq i \leq m$ there is an index $1 \leq r(i) < i$ such that the separator $S[i, r(i)]$ defined by $S[i, r(i)] = C[i] \cap (C[1] \cup ... \cup C[i-1])$ is contained in $C[i] \cap C[r(i)]$;

3. construct the undirected graph whose nodes are the cliques $C[1], ..., C[m]$, where for each $i = 1, ..., m$, the clique $C[i]$ is connected to the clique $C[r(i)]$ with an undirected edge if $S[i, r(i)] \neq \emptyset$.

Thus fast propagation procedures exist in high dimensional systems provided that the dimension of all cliques is small relative to the dimension of the state space.

In dynamic problems we can determine the junction tree $\mathcal{T}_T$ associated to the Bayesian probabilistic network $\mathcal{G}_T$ over the states $\Theta_T$ by applying the construction above. The cliques $C_T[1], ..., C_T[n_T]$ in $\mathcal{T}_T$ contain components of the state space $\Theta_T$ and the joint probability distribution $p(\Theta_T)$ is decomposable and can be written as

$$p(\Theta_T) = \frac{\prod_{i=1}^{n_T} p^{(i)}(\Theta_T[i])}{\prod_{i=2}^{n_T} q^{(i)}(\Theta_T[i, r(i)])}. \quad (1)$$

Here $\Theta_T[i]$ is a subvector of $\Theta_T$ whose components lie in the clique $C[i], 1 \leq i \leq n_T$. The vector $\Theta_T[i, r(i)]$ is a subvector of $\Theta_T[i]$ and $\Theta_T[r(i)]$ for $r(i) < i$ and its components are in the separator $S_T[i, r(i)]$ contained in the linked cliques $C_T[i]$ and $C_T[r(i)]$ for the running intersection property. Finally $p^{(i)}$ and $q^{(i)}$ are the marginal densities of $\Theta_t[i]$ and $\Theta_T[i, r(i)]$ respectively $1 \leq i \leq n_T$.

For instance all Markov processes have junction trees whose nodes all lie on a single line. But a wide range of other more complicated dependence structures can be represented in terms of junction tree, see, for example, Spiegelhalter et al. (1993), Goldstein (1993). Smith and Papamichail (1999) discuss in detail the formal structure of such transformation from dynamic DAG's to dynamic junction trees.

When all the variables in the dynamic system are Gaussian, fast propagation algorithms over dynamic junction trees are well known, see for example Smith et al. (1995). The probability associated to each clique is sequentially updated in the light of incoming data via Kalman filtering (West and Harrison, 1997) and calculations are all in closed form as it is shown in section 2. In the non-linear case, when the sampling distribution of the observations is not Gaussian, then typically the posterior distribution of the states in the system domain can not be determined in closed form and its calculation usually becomes intractable. However the conditional independence structure implicit in the graph still remains valid.

In section 3 we shall present an algebraic approximate procedure that updates the probability distribution in each clique for non-linear dynamic processes. By applying the approximate updating procedure proposed by West et al. (1985) for the estimation of the posterior probability of dynamic generalised linear models we shall be able to develop a propagation algorithm for dynamic junction trees. This algebraic algorithm has the advantage of propagating very quickly the information through complicated dynamic processes. The efficiency of this approximation is checked in section 4 by using the Hellinger metric. In section 5 the application of such a propagation procedure is discussed for the Poisson dynamic system associated to a gaseous mass release process.

## 2 GAUSSIAN JUNCTION TREES

Let $Y_t$ denote an n-dimensional vector of observations at time $t$ for $t = 1, ..., T$. The time series $Y_t$ is a realisation of a dynamic process whose state space at time $t$ is denoted by $\Theta_t = (\theta_1, ..., \theta_t)$. The class of stochastic processes we are interested in can be characterised by the following properties:

(i) given the states $\Theta_T = (\theta_1, \theta_2, ..., \theta_T)$ the variables $Y = \{Y_1, ..., Y_T\}$ are assumed all independent of each other i.e. $\perp\!\!\!\perp_{t=1}^T Y_t | \Theta_T$.

(ii) The vector $Y_t, t = 1, ..., T$ can be partitioned in $n_t$ subvectors $Y_t = (Y_t[1], ..., Y_t[n_t])$ for $1 \leq n_t \leq n$ and the density or probability function of each $Y_t[j]$ for each $j = 1, ..., n_t$ depends only on a linear combination of the states $\theta_t[j]$, say

$$\lambda_t[j] = F_t[j]\theta_t[j], \qquad (2)$$

where $F_t[j]$ is a known matrix. Notice that, given $\lambda_t[j]$, the vector of observations $Y_t[j]$ is assumed independent of all the other components of $Y_t$.

(iii) At any time $t$ the joint density of the states $\Theta_t$ is Markov with respect to a decomposable directed graph $\mathcal{G}_t$ whose set of nodes contains the states $\Theta_t$. The states $\theta_t[j]$ defined in (ii) are assumed to lie in a clique, say $C_t[j]$, in $\mathcal{G}_t$.

An example of such a process is pictured in the graphs in Figure 1. Property (ii) indicates that observations at time $t$ only give direct information about components of the state vector at that time and assumption (iii) ensures that observation vectors give direct information about single cliques.

This situation arises very often. For example in a Dynamic Linear Model (see e.g. West and Harrison, 1997 and Harrison and Stevens, 1976) conditional on the values of states at time $t$, the observations $Y_1, ..., Y_T$ are independent and furthermore $Y_t$ only depends on current states which lie in the same clique. In the spatio-temporal process described in Smith et al. (1995) an observation is taken at time $t$ whose expectation is linear in states, the linear combination depending upon the site at which the observation is taken.

The propagation procedures to update the Gaussian probability density $p(\Theta_t)$ over the states $\Theta_t$ in the light of the observation vector $Y_t$ are reasonably well-known and can be defined as follows. Notice that the same procedure can be applied to determine how states should be updated from a time series of observations $\{Y_1, ..., Y_T\}$, because the algorithm below can be simply iterated through the time sequence for $t = 1, ..., T$.

Firstly construct the junction tree $\mathcal{T}_t$ from the DAG $\mathcal{G}_t$ and let $C_t[1], ..., C_t[n_t]$ denote the cliques in $\mathcal{T}_t$ for some $t$. Notice that from the assumptions above each clique $C_t[j]$ contains the states $\theta_t[j]$ that are related to some observation vector $Y_t[j]$ via the linear combination in (2). Therefore we assume that observations $Y_t[j]$ give direct information about one clique $C_t[j]$ for $j = 1, ..., n_t$.

Suppose that the states $\theta_t[j]$ in clique $C_t[j]$ have a Gaussian density with mean $\mu_t[j]$ and covariance matrix $\Sigma_t[j]$ and a Gaussian vector $Y_t[j]$ is observed for which $Y_t[j] \perp\!\!\!\perp \theta_t[j] | \lambda_t[j]$. Let $m_t[j] = F_t^T[j]\mu_t[j]$ and the matrix $W_t[j]$ be the prior mean and covariance matrix of $\lambda_t[j] = F_t^T[j]\theta_t[j]$ at time $t$. After observing $Y_t[j]$, the posterior distribution of $\lambda_t[j]$ given $Y_t[j]$ can be calculated as Gaussian with posterior mean $m_t^*[j]$ and posterior variance $W_t^*[j]$ obtained by standard conjugate analysis. Known results from multivariate normal theory (see for instance Mardia et al., 1979) now establish that, provided $(Y_t[j], \theta_t[j])$ are jointly Gaussian, then the posterior density of $p(\theta_t[j]|Y_t[j])$ is Gaussian with mean $\mu_t^*[j]$ and covariance matrix $\Sigma_t^*[j]$ calculated as

$$\mu_t^*[j] = \mu_t[j] + A_t[j](m_t^*[j] - m_t[j])$$

$$\Sigma_t^*[j] = \Sigma_t[j] + A_t[j](W_t^*[j] - W_t[j])A_t^T[j]$$

where $A_t[j] = \Sigma_t[j]F_t[j](F_t^T[j]\Sigma_t[j]F_t[j])^{-1}$.

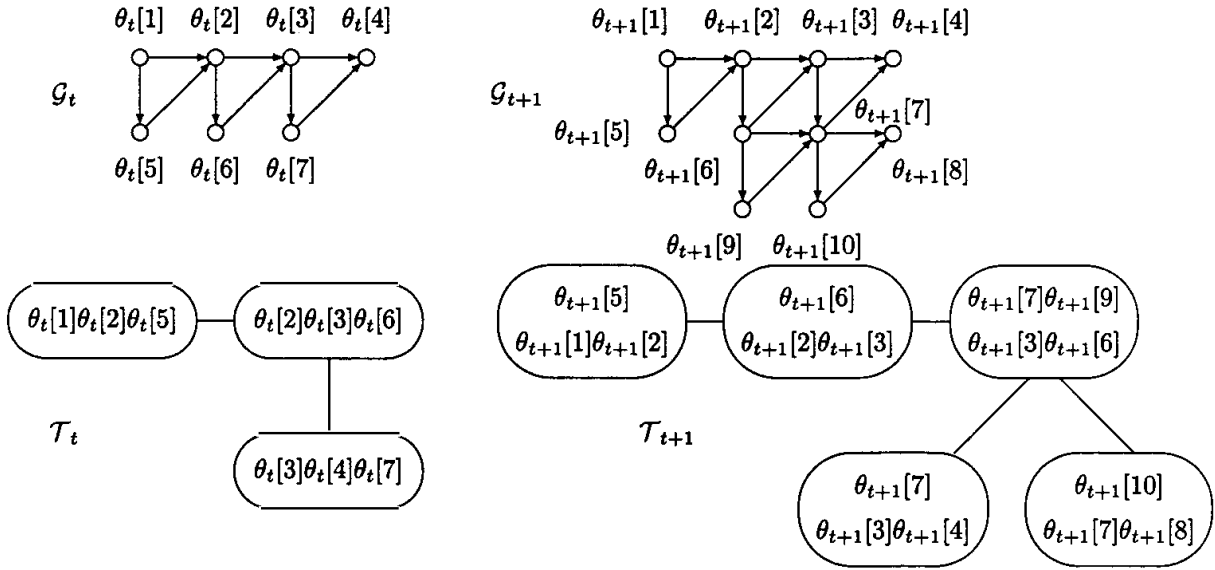

Figure 1: Example of Bayesian networks and junctions trees for a dynamic process at time $t$ and $t+1$

So the posterior density of $\theta_t[j]$ of the random vector associated with the clique $C_t[j]$ conditional on $Y_t[j]$ is easy to calculate. Henceforth we shall call these densities partial marginals. Of course what we really need is the posterior clique margins given all the data. Once these have been calculated, the information acquired by the updated clique(s) must be transmitted to the rest of the system. So we need to define a propagation algorithm which updates the probability density of the cliques that have not directly obtained information from the data and adjusts the full probability distribution of the system in the light of all the data. Notice that the probability decomposition of $p(\Theta_t)$ given by the junction tree $\mathcal{T}_t$ is still valid a posteriori. Algorithms that transmit information from clique margins to the all system have been around for some time for discrete probability distributions (see Spiegelhalter et al. (1993), Dawid (1992), Jensen et al. 1994) and are now well documented (see for example Almond (1995), Jensen (1996)). Gaussian analogues of these procedures are very straightforward and are developed in Lauritzen (1992) and Smith et al. (1995). One propagation algorithm, which is based on a two-step algorithm described in Jensen (1996), is outlined below.

We first introduce some notation. In the junction tree $\mathcal{T}_t$, for $t \leq T$, the *neighbours* of a clique $C_t[j], 1 \leq j \leq n_t$ are the cliques $C_t[k]$ for which either $r(j) = k$ and $S[j, r(j)]$ is a separator or $r(k) = j$ and $S[k, r(k)]$ is a separator in $\mathcal{T}_t$. Information can only be propagated through neighbouring cliques.

To simplify the notation, let $U, V$ denote two neighbouring cliques in $\mathcal{T}_t$, with separator $S = U \cap V$. The cliques $U, V$ have distribution with respective mean vectors $(\mu_{U \setminus S}, \mu_S)$ and $(\mu'_{V \setminus S}, \mu'_S)$ and covariance matrices

$$\begin{bmatrix} \Sigma_{U \setminus S} & \\ \Sigma_{U \setminus S, S} & \Sigma_S \end{bmatrix} \begin{bmatrix} \Sigma'_{V \setminus S} & \\ \Sigma'_{V \setminus S, S} & \Sigma'_S \end{bmatrix}$$

We shall say the the clique $V$ *absorbs* from a neighbouring clique $U$ with common separator $S$, if the updated distribution of $V$ has new mean vector $(\mu''_{V \setminus S}, \mu''_S)$ and covariance matrix

$$\begin{bmatrix} \Sigma''_{V \setminus S} & \\ \Sigma''_{V \setminus S, S} & \Sigma''_S \end{bmatrix} \text{ where } \mu''_S = \mu_S, \ \Sigma''_S = \Sigma_S$$

calculated as

$$\mu''_{V \setminus S} = \mu'_{V \setminus S} + A_{V \setminus S}(\mu_S - \mu'_S)$$
$$\Sigma''_{V \setminus S} = \Sigma'_{V \setminus S} + A_{V \setminus S}(\Sigma_S - \Sigma'_S)A^T_{V \setminus S} \quad (3)$$
$$\Sigma''_{V \setminus S, S} = A_{V \setminus S}\Sigma_S$$

where $A_{V \setminus S} = \Sigma'_{V \setminus S, S}(\Sigma'_S)^{-1}$.

Thus to transmit information from the updated margins of the observed cliques we follow the two step-algorithm proposed by Jensen et al. (1994) defined by the sequence of operations below:

(i) choose a clique in the junction tree $\mathcal{T}_t$, and name it *root*.

(ii) apply "absorb" from the cliques that have received information to neighbouring cliques following paths towards the root clique, until the root is reached (collect evidence);

(iii) apply "absorb" from the root to the neighbouring cliques following paths that departs away from the

root, until all the cliques have "absorbed" information (distribute evidence).

At this stage the derived mean vectors and covariance matrices of the clique $C_t[j]$ will be those of $\Theta_t[j]$ conditional on all data $\mathbf{Y}_t$ (a proof of this assertion can be found in Dawid, 1992). The full joint probability density conditional on the data can now be obtained from the decomposition in equation (1).

Thus provided $(\Theta_T, \mathbf{Y}_T)$ are jointly Gaussian, the posterior distribution $p(\Theta_T|\mathbf{Y}_T)$ is simple to calculate algebraically. Indeed it is Gaussian with mean and covariance matrix calculated in closed form as sequence of operations given in (3). However, unless the distribution of $Y_t[j]$ given $\lambda_t[j]$ is purely discrete or Gaussian, there are only few distributions for which an exact algebraic algorithm like the one given above exists and for which the vector of states $\Theta_T$ continues to lie in a recognised family of distributions, see Lauritzen (1992) for some exceptions. The problem is that although the probability breakdown in equation (1) is still valid, the posterior marginal densities of the states in cliques that receive information can no longer be written in closed form. To sidestep these problems Goldstein (1993) works in a simpler context of linear estimators and produces algebraic algorithms to update the probability distribution over the states. An alternative algorithm (Thomas et al., 1992) uses numerical integration methods such as Markov Chain Monte Carlo methods to update the clique margins. This methods, which can calculate numerical distributions to arbitrary degrees of accuracy, has much to recommend it. However because its output is not algebraic it tends to be very slow in complex and very large problems. Furthermore it is much more difficult to investigate how the system learns from the data and to monitor the assumed dependence structures in the light of the data.

In this paper we suggest a different route. This uses the updating techniques of Dynamic Generalised Linear Models D.G.L.M's (West and Harrison, 1997). We treat West and Harrison's algorithm as if it were a dynamic approximation technique of a full Bayesian analysis, This interpretation to their algorithm is developed in Smith (1992). Henceforth for the sake of simplicity we shall assume that $\lambda_t[j] = F_t^T \theta_t[j]$ is one-dimensional. However we remark that multivariate analogues are straightforward to develop.

## 3  DYNAMIC GENERALISED LINEAR JUNCTION TREES

Suppose a random variable $\eta_t[j]$ belongs to some parametrised family of densities $\Psi$ which is closed under sampling of an observation $Y_t[j]$ whose distribution conditional on $\eta_t[j]$ lies in a family $\Upsilon$. For simplicity of exposition West and Harrison (1989) choose to restrict $\Upsilon$ to be the exponential family although this condition is not strictly necessary for their algorithm to work. Now assume

$$\lambda_t[j] = g_j(\eta_t[j]) \qquad (4)$$

where $g_j$ is a known invertible function and $\lambda_t[j]$ is a linear function of the uncertain state vector $\theta_t[j]$ as in equation (2). Because the distribution of $\lambda_t[j]$ is Gaussian it is unlikely that the density of $\eta_t[j] = g_j^{-1}(\lambda_t[j])$ will lie in $\Psi$. However, provided that the function $g_j$ is chosen appropriately and $\Psi$ is two dimensional it will often be possible to find a density $\hat{p}(\eta_t[j]) \in \Psi$ which is very close to the transformed Gaussian density $p(\eta_t[j])$ of $\eta_t[j]$. In fact in practice this will often be true when for each index $j$ the functions $g_j$ are identity maps - see the example in section 5. A more exact but complicated choice for $g_j$ specified as a function of the predictive distribution function of $\eta_t[j]$ is discussed in Smith (1992).

In practice, to ensure not only that recurrence equations can be written in a closed form, but also that the states of the process retain relationships between each other which are straightforward to explain, we usually choose simple functions $g_j$ and elementary ways of approximating $p(\eta_t[j])$ by $\hat{p}(\eta_t[j])$, for example, by equating the moments of these two distributions.

Conversely it should be possible to find a transformed Gaussian density $\hat{q}(\eta_t[j])$ which is close to any density $q(\eta_t[j]) \in \Psi$ that might arise in our analysis. Appropriate definitions of closeness will be discussed in the next section. So assume that a good simple approximation of this type is available. We can now find an approximate algebraic method of probability propagation analogous to the one defined in section 2 for Gaussian domains.

Suppose that $Y_t[j]$ depends on $\eta_t[j] = g_j^{-1}(\lambda_t[j])$ where $\lambda_t[j] = F_t^T[j]\theta_t[j]$ is a linear function of $\theta_t[j]$ whose components all lie in the same clique $C_t[j]$. If $\theta_t$ is Gaussian then in particular $\lambda_t[j]$ will be Gaussian. Now identify the density $\hat{p}(\eta_t[j])$ lying in a family closed under sampling to $Y_t[j]|\eta_t[j]$ which approximates the density $p(\eta_t[j])$.

Next perform a conjugate analysis to find the posterior density $\hat{p}(\eta_t[j]|Y_t[j])$ which will lie in $\Psi$. Identify the Gaussian density on $\lambda_t[j]$ for which $g_j^{-1}(\lambda_t[j])$ has approximately the density $\hat{p}(\eta_t[j]|Y_t[j])$. Notice that, through this operation, we have approximately updated the marginal distributions of the states in the cliques $C_t[j]$ that are "observed" and furthermore our approximation sets each of these distribution to be

| Prior Distribution of $\eta_t[j], \lambda_t[j]$ | Distribution of $Y_t[j]\|\eta_t[j]$ | Posterior mean $m_t^*[j]$ of $\lambda_t[j]$ | Posterior variance $w_t^{*2}[j]$ of $\lambda_t[j]$ |
|---|---|---|---|
| Normal $\eta_t[j] \sim N(m_t[j], w_t^2[j])$ $\lambda_t[j] = \eta_t[j]$ | Normal $N(\eta_t[j], V_t[j])$ | $(1 - A_t[j])m_t[j] + A_t[j]y_t[j]$ | $A_t[j]V_t[j] =$ $(1 - A_t[j])w_t^2[j]$ $A_t[j] = \dfrac{w_t^2[j]}{w_t^2[j] + V_t[j]}$ |
| Gamma $\eta_t[j] \sim G(\alpha_t[j], \beta_t[j])$ $\alpha_t[j] = m_t^2[j]/w_t^2[j],$ $\beta_t[j] = m_t[j]/w_t^2[j]$ $\lambda \sim N(m_t[j], w_t^2[j])$ | Poisson $Po(\eta_t[j])$ | $(1 - A_t[j])m_t[j] + A_t[j]y_t[j]$ | $A_t[j]m_t^*[j]$ $A_t[j] = \dfrac{w_t^2[j]}{(w_t^2[j] + m_t[j])}$ |
| Log-normal $\log \eta \sim N(\alpha_t[j], \beta_t[j])$ $m_t[j] = E(\eta_t[j]) = e^{(\alpha_t[j] + \frac{1}{2}\beta_t[j])}$ $w_t^2[j] = Var(\eta_t[j]) = (e^{\beta_t[j]} - 1)m_t^2[j]$ $\lambda_t[j] \sim N(m_t[j], w_t^2[j])$ | Log-normal $\log Y_t[j]\|\eta_t[j] \sim$ $N(\log \eta, V_t[j])$ | $(m_t[j])^{(1-A_t[j])}(y_t[j])^{A_t[j]}$ $A_t[j] = \dfrac{\log(m_t^2[j] + w_t^2[j]) - \log m_t^2[j]}{\log(m_t^2[j] + w_t^2[j]) - \log m_t^2[j] + V_t[j]}$ | $(e^{A_t[j]V_t[j]} - 1)m_t^{*2}[j]$ |

Table 1: Some illustrative examples where the link function $g(\cdot)$ is the identity function. ($m_t[j]$ and $w_t^2[j]$ are the prior mean and variance of $\lambda_t[j]$), respectively.

Gaussian. Thus now we can transmit information and update the probabilities in the junction tree by using the propagation algorithm described in section 2 for Gaussian systems and so calculate the full posterior distribution over the states. In a time series context this forms the prior density of the states for the next vector of observations.

Assuming that the sampling distribution of the observations $Y_t[j]$ is Poisson or Log-normal and that the observations $Y_t[j]$ depend on a linear combination $\lambda_t[j]$ of states which lie in a single clique, Table 1 displays the expressions for the mean and variance of the posterior distribution of $\lambda_t[j]$ calculated by applying the approximate updating procedure described above, when the link function $g_j$ is the identity (so $\lambda_t[j] = \eta_t[j]$). Notice that in Table 1, the updating procedure essentially consists of approximating the posterior distribution of $\lambda_t[j]$ as a Gaussian with mean and variance calculated from the posterior distribution of $\eta_t[j]$.

Notice that the obvious difference from the Gaussian case is that now the posterior variance of $\lambda_t[j]\|Y_t[j]$ can be a function of $Y_t[j]$. Other examples of such updating using nonlinear link functions are given in West and Harrison (1997).

Hence the calculation of the marginal distribution over the "observed" cliques using equations like those in Table 1 and then using equation (2) to perform this approximate probability propagation is just as quick and simple as the exact case when the sampling distribution is Gaussian. The speed and simplicity arises because the method is algebraic and the approximating distribution over the states $\Theta_T$ is Gaussian. So a very slight change to the code allows the quick processing of data which does not have a Gaussian sampling distribution. This is now coded in software: see Smith and Faria (1997), Smith and Papamichail (1999). Of course the validity of the approximate updating algorithm described above depends critically on how well the true posterior density of $\Theta_t|\mathbf{Y}_t$ is approximated by the Gaussian one calculated by our algorithm. Methods to check the accuracy of this approximation are given in the following section.

## 4 MONITORING THE DYNAMIC APPROXIMATION

In this paper, we choose the Hellinger metric defined by

$$d_H(f, h) = \left(1 - \int f^{1/2}(x)h^{1/2}(x)dx\right)^{1/2}$$

as divergence measure between densities. It always takes values between zero and one and when $f$ and $h$ are absolutely continuous equals one only when the supports of $f$ and $h$ intersect on a set of measure zero. The Hellinger metric is topologically equivalent to the variation metric

$$d_V(f,h) = 1/2 \int |f(\mathbf{x})-h(\mathbf{x})|d\mathbf{x} = \sup_A |P_f(A)-P_h(A)|$$

where $A$ is any subset of values of $\mathbf{x}$ and $P_f$ and $P_h$ are respectively the probability of the set $A$ under $f$ and $h$. Explicitly it can be shown that

$$d_H^2(p,\hat{p}) \leq d_V((p,\hat{p}) \leq \sqrt{2}d_H(p,\hat{p})$$

see, for example Reiss (1989). So, in particular, when the Hellinger distance between two densities $f$ and $g$ is small then we know that all probability statements associated with the two different models are close. The Hellinger metric is therefore a useful measure of the closeness of two densities. Write

$$I(f,g) = 1 - d_H^2(f,g)$$

Then, for example, the Hellinger distance between two normal densities $f_1$ and $f_2$ with respective means and variances $(\mu_1, \sigma_1^2), (\mu_2, \sigma_2^2)$ can be calculated from

$$I^2(f_1, f_2) = \frac{2\sigma_1\sigma_2}{\sigma_1^2+\sigma_2^2} \exp\{-1/2(\sigma_1^2+\sigma_2^2)^{-1}(\mu_1-\mu_2)^2\}.$$

In fact a property of the Hellinger distance which makes it easier to manipulate than the variation metric is that $d_H^2(f,g)$ can be calculated in closed form for densities in most standard families, including the exponential family. It is also sometimes possible to explicitly write down the Hellinger distance between two densities from different families, see Smith (1995).

Thus when $f$ is Gaussian with mean $\mu$ and variance $\sigma^2$ and $f'$ is a Gamma density with the same mean and variance, then $I(f,f')$ after some algebra can be calculated as

$$I^2(f,f') = (2\pi)^{-1/2}2^{(\alpha-1)}\alpha^{1/2\alpha}\frac{(\Gamma(1/4[\alpha+1]))^2}{\Gamma(\alpha)}e^{1/2\alpha} \quad (5)$$

where $\alpha = \mu^2/\sigma^2$.

We note that the two properties listed below also hold true both for the variation metric and the popular Kullback-Leibler separation measure (Kjaerullf, 1992). Suppose that $p$ and $\hat{p}$ are joint densities on $X = (X_1, X_2)$ which have different margins $p_1$ and $\hat{p}_1$ on $X_1$ but whose conditional densities of $X_2|X_1$ agree. Then, directly from (4) we have that

$$d_H(p,\hat{p}) = d_H(p_1,\hat{p}_1) \quad (6)$$

Now the algorithm we suggest above approximates only the distribution of $\lambda$. The distribution of all the states given $\lambda$ is the same both for the true and the approximating density. It follows from (6) that the closeness of the full joint density over all states $\Theta_T$ depends only on the closeness of our approximation of the one dimensional normal posterior density of $\lambda$ to the true posterior density of $\lambda$. If we can ensure the approximation of the distribution of $\lambda$ is close to the true one under our algorithm then our approximation will be good in the sense that *all* approximate probability statements about states will be close to their exact analogues.

In a dynamic context predictions usually depend only on a subvector $\Theta_k$ of states $\Theta_T$. For example in the D.G.L.M predictions only depend on the current state vector. Similarly in Smith et al. (1995) the components of $\Theta_T$ lying in cliques which are not parents of other cliques are the only components required for the prediction of the future levels of contamination. Now it is easily checked that

$$d_H(p,\hat{p}) - d_H(p_k,\hat{p}_k) \geq 0 \quad (7)$$

where $p_k, \hat{p}_k$ are the densities of $\Theta_k$ under the true model and its approximation. Now consider accommodating a sequence of observations $Y_1, \ldots Y_t, \ldots$ each vector having components which are functions, possibly with error, of variables in a single clique, to obtain an approximate posterior density $\hat{p}$ on states. Although errors associated with the approximation, given in equation (8), will tend to accumulate in $\hat{p}$, this accumulation will tend to be offset by the reduction of error in (7) associated with the act of marginalising states relevant for linear prediction.

## 5 AN EXAMPLE FROM DISPERSAL MODELLING

In the Gaussian model given in Smith et al. (1995), used for predicting air concentrations after a nuclear accident, it is often most natural to assume that the distribution of observations conditional on their states is Poisson. The states $\Theta_T$ are the quantities of mass under puffs and fragments of contamination where puffs are emitted stochastically from a chimney and then directed by a known wind-field across space. The Markov nature of the stochastic emission process and deterministic fragmentation process, means that the joint distribution of mass fragments at any time is decomposable with its clique dimension not greater than six (see Smith et al., 1995, for more details)

The appropriate updating equations for the linear combination of state vectors $\lambda_t[j] = \mu_t[j]$ we learn about, is displayed in Table 1, given a model with identity link function. It is also possible to calculate an upper bound on how good the approximation is by using the Hellinger divergence measure $d_H$. Let $p_0(\Theta_T)$ denote the Gaussian prior density on the states $\Theta_T$. Let $p$ and $\hat{p}$ denote the posterior density on $\Theta_T$ given the true normalised Gamma likelihood $L_2$ associated with a Poisson observation $Y_t$ or a normalised Gaussian approximation $L_1$ of the D.G.L.M., respectively. Then by using the equality (6), the Cauchy-Schwartz inequality on $I(p,\hat{p})$ and the triangle inequality on $d_H$ it can be shown after tedious algebra that

$$d_H^2(\hat{p},p) \leq$$
$$1 - (1-\varepsilon_1)\left[\frac{\int p_0(\theta)\hat{L}_2(\theta)d\theta}{\int p_0(\theta)L_1(\theta)d\theta} + \frac{\varepsilon_2}{\int p_0(\theta)L_1(\theta)d\theta}\right]^{-1/2},$$

where

$$\varepsilon_1 = \sqrt{2}\tau[d_H(L_1,\hat{L}_2) + d_H(\hat{L}_2,L_2)]$$
$$\varepsilon_2 = c_0[(c_2-\hat{c}_2)^2 + 2c_2\hat{c}_2 d_H^2(k_2,\hat{k}_2)]^{1/2}$$

where $\hat{L}_2$ is a normalised Gaussian likelihood with the same mean and variance as $L_2$,

$$c_0^2 = \int p_0^2(\theta)d\theta, \quad c_2^2 = \int p_2^2(\theta)d\theta, \quad \hat{c}_2^2 = \int \hat{L}_2^2(\theta)d\theta,$$

$$k_2^2 = \frac{L_2^2}{c_2^2}, \hat{k}_2^2 = \frac{\hat{L}_2^2}{\hat{c}_2^2}$$

$$\tau = \frac{(\int 2p_0^2(\theta)L_1(\theta)d\theta)^{1/2}}{(\int p_0(\theta)L_1(\theta)d\theta)}.$$

It can be shown that $\varepsilon_1$ and $\varepsilon_2$ are very small when using the D.G.L.M approximation when the parameter $\alpha$ in the Gamma likelihood discussed in (5) is of moderate size. Note from (5) that $\varepsilon_1$ can be calculated explicitly. All terms in $\varepsilon_2$ can also be calculated explicitly, except for $d_H^2(k_2,\hat{k}_2)$. This term can be given a tight explicit upper bound using the triangle inequality (Gargoum, 1998). So we obtain an explicit upper bound for $d_H^2$ as a function of the parameters of the model and the observation.

These calculations are now implemented within RODOS software (Smith and Faria, 1997) and are used as a diagnostic to check whether the sequential Gaussian updating described above is actually theoretically justified. For the types of data we experience it has been checked through many runs of this algorithm that the Hellinger distances approximations between the approximate and actual densities are nearly always bounded by very small numbers.

An exception to this is when the data is incompatible with the probability statement of the model, however this extreme event is typically picked up by diagnostics run with the model. We also may have large discrepancies when the Poisson counts are small. In the chosen context, main concern is focused to cases where Poisson counts are large as they correspond to high levels of radioactivity. So the clique states $\Theta_k$ of interest, are usually geographically remote from low counts. It follows that the approximation of the distribution of $\Theta_k$ usually remains a good one. So we have shown that in at least one application, these methods are not only very quick but also surprisingly accurate.

In Settimi and Smith (1998) the goodness of such an algebraic approximation is analysed by using MCMC methods. The approximated posterior distribution of the states for a dynamic process with Poisson sampling distribution is obtained from the algebraic procedure and compared with the "true" posterior distribution of the states calculated numerically by a Gibbs sampler. The results show that the algebraic approximation is very good when observations are consistent with the assumed model, however a decay in the efficiency of the approximation is noted as the observations tend to depart from the predicted values of the model. To be more precise, when unexpectedly low counts are observed, the algebraic method seems to overestimate future observations, while, if large counts occur, the approximate method gives prediction similar to the Gibbs sampler output and the effect of such extreme data is negligible.

Using the rather crude bounds on the Hellinger distance, it was possible to show that the aggregated effect of the sequence of approximation used within the method could distort any probability forecast by no more than 0.01. Numerical investigation currently being undertaken appear to show that this bound is of order of magnitude too large for the predictions of high levels of contamination. However even aggregating on the possible distortions ignoring the effect (6) and using these crude bounds within the context in which the software is implemented, we notice that the contribution of error associated with the approximation is confounded by other sources of error associated with the model (see Ranyard and Smith, 1997, for a discussion of these modelling issues).

## 6   CONCLUSION

This analogue of dynamic generalised linear models, when used on junction trees, gives a quick computational approach for dealing with non-normal data which is easy to understand, gives a closed form updating algorithm and provides an approximation whose validity can be checked numerically - for example by using the Hellinger distance metric. In an iterative system where quick calculation is essential and easy interpretation is paramount it is our opinion that the methods described in this paper provide a practical methodology for quick Bayesian inference in complex dynamic systems.

## Acknowledgements

The research work of Raffaella Settimi and Jim Q. Smith was supported by Engineering and Physical Sciences Research Council (grant no. GR/K72254)

## References

Almond, R.G. (1985). *Graphical Belief Modeling* Chapman & Hall: New York.

Boyen, X. and Koller, D. (1998). Tractable inference for complex stochastic processes. In *Proceedings of the Fourteenth Conference in Uncertainty in Artificial Intelligence.* Morgan Kauffman, San Francisco, CA, 33–42.

Dawid A. P. (1992). Applications of a general propagation algorithm for probabilistic expert systems. *Statistics and Computing*, 2, 25–36.

Durbin, J. and Koopman, S.J. (1999). Time series analysis of non-Gaussian observations based on state space models from both classical and Bayesian perspectives (with discussion). To appear in *Journal of Royal Statistical Society*, **B**.

Gargoum, A. S. (1998). Issues in Bayesian forecasting of dispersal after a nuclear accident. Ph.D. thesis, Statistics Department, University of Warwick, UK.

Goldstein, M. (1993). Prediction under the influence: Bayes linear influence diagrams for prediction in a large brewery. *The Statistician*, 42, 445–459.

Harrison, P. J. and Stevens, C. F. (1976). Bayesian forecasting (with discussion), *Journal of the Royal Statistical Society*, **B**, 38, 203–247.

Jensen, F., Jensen, F.V. and Dittmer, S.L. (1994). From influence diagrams to junction trees. In *Proceedings of the Tenth Conference in Uncertainty in Artificial Intelligence.* Morgan Kauffman, San Francisco, CA, 367–373.

Jensen, F.V. (1996). *An introduction to Bayesian networks.* U.C.L. Press: London.

Kjaerulff, U. (1992). A computational scheme for reasoning in dynamic probabilistic networks. In *Proceedings of the Eighth Conference in Uncertainty in Artificial Intelligence* Morgan Kaufmann, San Mateo, CA, 121–129.

Lauritzen, S. L. (1992). Propagation of probabilities, means and variances in mixed graphical association models. *Journal of American Statistical Association*, 87, 1098–1108.

Mardia, K.V., Kent, J.T. and Bibby, J.M. (1979). *Multivariate analysis.* Academic Press: London.

Ranyard, D.C. and Smith, J.Q. (1997). Building a Bayesian model in scientific environment: managing uncertainty after an accident. In *The practice of Bayesian analysis* (S. French and J.Q. Smith, eds.), Arnold : London, 245–258.

Reiss, R. D. (1989). *Approximation Distribution of Order Statistics: with applications to nonparametric statistics.* New York: Springer-Verlag.

Settimi, R. and Smith, J.Q. (1998). A comparison of approximate Bayesian forecasting methods for non Gaussian time series. To appear in *Journal of Forecasting.*

Shepard, N. and Pitt, M.K. (1999). Analytic convergence rates and parametrisation issues for Gibbs sampler applied to state space models. To appear in *Journal of Time Series.*

Smith, J. Q. (1992). A comparison of the characteristics of some Bayesian forecasting models. *International Statistical Review*, 1, 75-87.

Smith, J. Q. (1995). Approximating Bayesian models using the Hellinger metric. University of Warwick, Research Report.

Smith, J. Q. and Faria, A.E. (1997). Bayes networks for uncertainty handling in nuclear risk. In *Proceedings of the sixth topical meeting on Emergency preparedness and response,* American Nuclear Society, San Francisco, April 1997.

Smith, J. Q., French, S. and Ranyard, D. (1995). An efficient graphical algorithm for updating dispersal estimates of the dispersal of gaseous waste after an accident release. In *Probabilistic reasoning and Bayesian belief networks,* A. Gammerman (ed.). Alfred Walker, 125–140.

Smith J.Q. and Papamichail K.N. (1999). Fast Bayes and the dynamic junction forest. *Artificial Intelligence*, **107**, 99–121.

Spiegelhalter, D. J., Dawid, A. P., Lauritzen, S. L. and Cowell, R. G. (1993). Bayesian analysis in expert systems. *Statistical Science*, 8, 219-246.

Thomas, A., Spiegelhalter, D. J. and Gilks, W. R. (1992). BUGS: A program to perform Bayesian inference using Gibbs sampling. In *Bayesian Statistics 4*, J. M. Bernardo, J. O. Berger, A. P. David, and A. F. M. Smith (eds.), Clarendor press: Oxford, 837–42.

West, M. and Harrison, P. J. (1997). *Bayesian Forecasting and Dynamic Models.* 2nd edition, New York: Springer-Verlag.

West, M., Harrison, P. J. and Migon, H. S. (1985). Dynamic generalised linear models and Bayesian forecasting (with discussion). *Journal of American Statistical Association*, 80, 73–97.